\documentclass{article}

\newcommand{\final}{1}

\usepackage[final]{nips_2018}

\usepackage{natbib}
\usepackage[utf8]{inputenc} 
\usepackage[T1]{fontenc}    
\usepackage{hyperref}       
\usepackage{url}            
\usepackage{booktabs}       
\usepackage{amsfonts}       
\usepackage{nicefrac}       
\usepackage{microtype}      
\usepackage{graphicx}
\usepackage{color}
\usepackage{ifthen}
\usepackage{float}
\usepackage{amsmath}
\usepackage{amssymb}
\usepackage{amsthm}
\usepackage{wrapfig}
\usepackage{subfig} 
\usepackage{multirow}
\usepackage{algorithm}
\usepackage{algorithmic}

\newcommand{\Caption}[2]{\caption[#1]{{\footnotesize #1} {\footnotesize #2}}}

\definecolor{DeltaColor}{rgb}{0.039,0.73,0.71}
\definecolor{SigmaColor}{rgb}{0.98,0.45,0.0}
\definecolor{JediColor}{rgb}{0.8,0,0}
\definecolor{AlphaColor}{rgb}{0,0,0.8}
\definecolor{BetaColor}{rgb}{0.8,0,0.8}
\definecolor{GammaColor}{rgb}{0.514,0.34,0.224}
\definecolor{EpsilonColor}{rgb}{0.353,0.725,0.906}
\newcommand{\weikai}[1]{{\color{JediColor} Weikai: #1 $\qed$}}

\newcommand{\note}[1]{{\it\color{blue} #1}}
\newcommand{\nothing}[1]{}

\definecolor{AudioColor}{rgb}{0.56,0.34,0.62}

\definecolor{DeadlineColor}{rgb}{0.9,0.4,0} 

\definecolor{figred}{rgb}{1,0,0}
\definecolor{figgreen}{rgb}{0,0.6,0}
\definecolor{figblue}{rgb}{0,0,1}
\definecolor{figpink}{rgb}{1,0.63,0.63}

\newcounter{pccount}
\floatstyle{plain}

\newcommand{\filename}[1]{\url{#1}}
\newcommand{\foldername}[1]{\url{#1}}

\graphicspath{
	{figs/raster/}
	{figs/handdrawn/}
}

\title{Deep RBFNet: Point Cloud Feature Learning using Radial Basis Functions}

\author{
	Weikai Chen$^1$, Xiaoguang Han$^2$, Guanbin Li$^3$, Chao Chen$^3$, Jun Xing$^1$, 
	Yajie Zhao$^1$, Hao Li$^{1,4}$ \\
	$^1$USC Institute for Creative Technologies \\
	$^2$The Chinese University of Hong Kong, Shenzhen \\
	$^3$Sun Yat-sen University \\
	$^4$University of Southern California 
}

\nothing{
\author{
	Weikai Chen\thanks{Use footnote for providing further
		information about author (webpage, alternative
		address)---\emph{not} for acknowledging funding agencies.} \\
	USC Institute for Creative Technologies\\
	Los Angeles, CA\\
	\And
	Xiaoguang Han \\
	The Chinese University of Hong Kong, Shenzhen\\
	Shenzhen, China\\
	\AND
	Guanbin Li \\
	Sun Yat-sen University\\
	Guangzhou, China\\
	\And
	Chao Chen \\
	Sun Yat-sen University\\
	Guangzhou, China\\
}
}

\begin{document}
	
\maketitle

\begin{abstract}

Three-dimensional object recognition has recently achieved great progress thanks to the development of effective point cloud-based learning frameworks, such as PointNet~\cite{charles2017pointnet} and its extensions. However, existing methods rely heavily on fully connected layers, which introduce a significant amount of parameters, making the network harder to train and prone to overfitting problems.
In this paper, we propose a simple yet effective framework for point set feature learning by leveraging a nonlinear activation layer encoded by Radial Basis Function (RBF) kernels.
Unlike PointNet variants, that fail to recognize local point patterns, our approach explicitly models the spatial distribution of point clouds by aggregating features from sparsely distributed RBF kernels.
A typical RBF kernel, e.g. Gaussian function, naturally penalizes long-distance response and is only activated by neighboring points.
Such localized response generates highly discriminative features given different point distributions.
In addition, our framework allows joint optimization of kernel distribution and its receptive field, automatically evolving kernel configurations in an end-to-end manner.
We demonstrate that the proposed network with a single RBF layer can outperform the state-of-the-art Pointnet++ \cite{qi2017pointnet++} in terms of classification accuracy for 3D object recognition tasks.
Moreover, the introduction of nonlinear mappings significantly reduces the number of network parameters and computational cost, enabling significantly faster training and a deployable point cloud recognition solution on portable devices with limited resources.

\end{abstract}

\section{Introduction}
\label{sec:intro}

Despite the dramatic advancement of deep learning techniques for image analysis, the reasoning about 3D geometric data remains largely unexplored.
In 3D computer graphics and solid modeling, polygonal mesh remains the most popular form of three-dimensional content representation.
However, the combinatorial irregularity and complexity of meshes have led to the nontrivial difficulty in developing an efficient learning framework, especially tailored to such data structure.
Unlike mesh representations, point clouds are a simple and unified structure that offers great flexibility in terms of topological invariance.
In addition, the rapid democratization of 3D sensors has rendered point clouds one of the most important and convenient data sources for high-level semantic understanding for a variety of object recognition tasks.

The major challenge of learning point cloud features lies in its spatial irregularity and invariance to permutations, which make it infeasible for applying existing frameworks like convolutional neural networks (CNNs).
Some early works attempt to rasterize 3D volumes into regular voxel grids in order to utilize 3D CNNs.
However, such data transformation often leads to redundant computation and suffers from quantization artifacts that may interfere with the natural data invariance.
The recently proposed PointNet~\cite{charles2017pointnet} pioneers in directly applying deep learning to point clouds.
Through a series of feature transforms that are composed of fully connected layers, PointNet implicitly learns high-dimensional feature representations of points in an isolated manner. Although the subsequent pooling operation helps collecting a global signature for the input point cloud, it still fails to capture local features by its design.
The recently presented PointNet++ \cite{qi2017pointnet++} addresses this issue by recursively applying PointNet in a hierarchical fashion.
Nonetheless, the spatial distribution of the input point cloud is still not explicitly modeled by the proposed sampling and grouping paradigm.
Though PointNet and Pointnet++ have achieved impressive performances for 3D point cloud recognition, the costly use of fully connected layers yields a considerable amount of parameters, making the network cumbersome to train and prone to overfitting.
Furthermore, it remains unclear how to interpret the mechanism of MLPs and how they contribute to the extraction of meaningful features directly from coordinates of unordered points.

\begin{wrapfigure}{r}{0.30\textwidth}
	\begin{center}
		\includegraphics[width=0.30\textwidth]{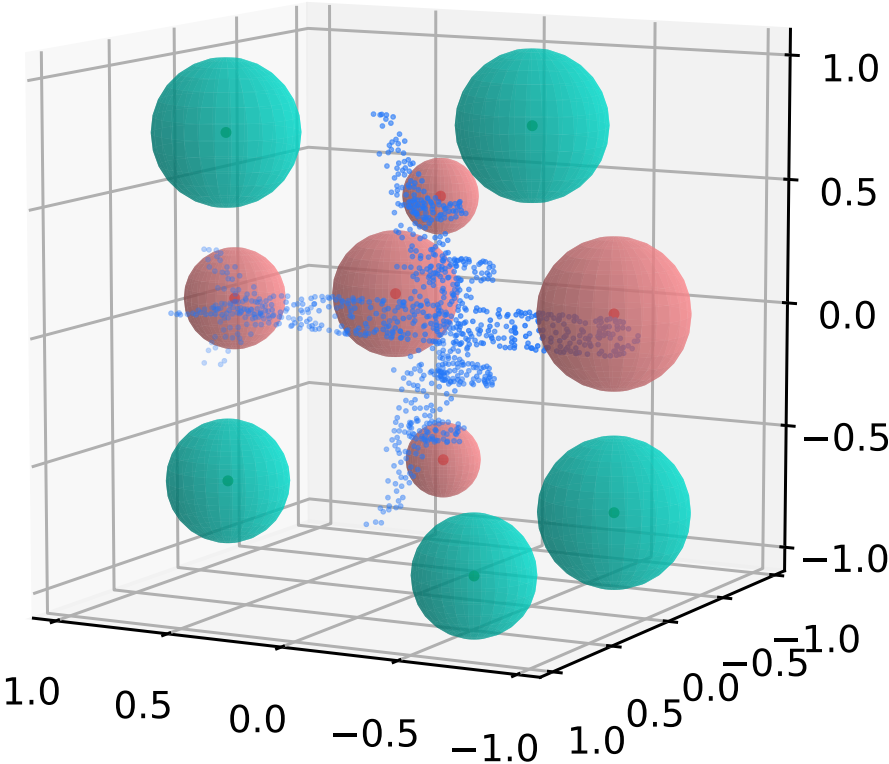}
	\end{center}
	\caption{Spatial encoding by the activations of RBF kernels. The activated kernels and its receptive field (represented as sphere) are marked in red; otherwise in green.}
	\label{fig:rbf}
\end{wrapfigure}
We address the above issues by proposing a novel framework that leverages the nonlinear activation encoded by RBF kernels.
Our approach is inspired by surface reconstruction approaches in computer graphics where a 2-manifold surface can be faithfully represented by a sparse combination of RBF kernels~\cite{carr2001reconstruction}.
Based on this paradigm, unstructured 3D point clouds can be well modeled by a sparse yet discriminative RBF-based representation, which can be learned effectively through our DeepRBFNet.
In particular, instead of directly passing point coordinates to fully connected layers, we first compute the response of each point with respect to a sparse set of RBF kernels scattered in 3D space.
The nonlinear activation modulated by RBF kernels brings several benefits.
First, it enables \textit{explicit modeling of the spatial distribution} of point clouds.
Due to the localized response, a typical Gaussian-based RBF kernel can only be activated by points within its receptive field.
Therefore, as illustrated in Figure~\ref{fig:rbf}, the activated pattern of RBF kernels can faithfully capture the spatial point distribution.
Furthermore, both the kernel center and the receptive field can be adaptively adjusted during optimization, generating highly discriminative, yet robust features given different input point clouds.
Second, thanks to the effective feature extraction, our approach outperforms the current state-of-the-art solutions in terms of performance even with \textit{significantly less computational cost and inference time}, which enables a deployable solution for hardware with limited resources, such as mobile or embedded devices.
Finally, our network is significantly \textit{more robust to noises and incomplete data} compared to other techniques thanks to the sparseness of kernel representation.

Our proposed solution is specifically tailored for learning point cloud features, which is general and independent of any existing structures.
As the response of an RBF kernel only depends on the distance from its origin, our network is invariant to point permutations.
Furthermore, our approach is flexible in considering different kernel functions, by combining which one could lead to more informative features.
Fusing output RBF convolution results of multiple kernels with point-wise MLP can further boost the performance.
Experimental results demonstrate that the network with a single layer of RBF convolution achieves better classification accuracy than state-of-the-art approaches on challenging 3D point cloud benchmarks. Moreover, our framework is capable of achieving much faster training, due to the simplicity of its architecture and the effectiveness of its feature extraction capabilities.

\section{Related Work}
\label{sec:related}
\paragraph{Shape Descriptor using Distance Distribution.}
Distance distribution is one of the most common handcrafted descriptors for characterizing global information of 3D shapes. It is first employed in~\cite{osada2002shape} to define a similarity metric between two shapes. 
Hamza and Krim~\cite{hamza2003geodesic} further apply geodesic distance for 3D shape classification, making it possible to capture pose-invariant features. The inner-distance was proposed in~\cite{ling2007shape} to build shape descriptors that are robust to articulation but discriminative for part structures. 
In our paper, the RBF kernels can be viewed as probing centers scattered in the 3D space. By computing the response of each point with respect to the kernel center, the point coordinates can be transformed to a nonlinear activation that solely depends on the distance distribution between points and kernel centers. 
The resulting features are highly discriminative and thus enable our algorithm to outperform state-of-the-art methods even with a relatively shallow network architecture. 
\vspace{-3mm}

\paragraph{RBF Network.}
Radial basis function network (RBFN) is a specific type of neural network that uses radial basis functions as activation functions. It was firstly proposed by Broomhead and Lowe ~\cite{broomhead1988radial} around three decades ago but has received very few attention in recent years.
We refer the readers to \cite{orr1996introduction} and \cite{ali2018radial} for a fundamental introduction. 
Despite the fact that RBFN has been applied in a variety of areas, e.g. face recognition~\cite{er2002face}, function approximation~\cite{wu2012using} and time series analysis~\cite{jones1990function}, it remains a virgin land to leverage RBFN for analyzing any form of geometric data.
In fact, in computer graphics, RBF has demonstrated its strong competency for reconstructing a 3D surface~\cite{carr2001reconstruction}.
Our work is motivated by the fact that via carefully optimizing the center position and kernel size, a set of RBF kernels can faithfully reconstruct smooth, manifold surface from non-uniformly sampled point cloud data.
The distance metric (Euclidean distance) for measuring spatial point distribution is particularly suitable for RBF kernel to encode and learn from.
We are the first to integrate RBFN with current deep learning framework and prove its efficiency for feature representation learning on point clouds. 
\vspace{-3mm}

\paragraph{Deep Learning on 3D Data.}
The 3D data can be represented in a variety of forms, leading to multiple lines for 3D deep learning. 
{\it Volumetric CNNs}:
As the pioneers in volumetric learning, \cite{wu20153d} and \cite{maturana2015voxnet} first propose to apply 3D CNNs on the voxelized content. 
However, voxelization based approach suffers from loss of resolution and high computational cost of 3D convolutions.
FPNN \cite{li2016fpnn} converts 3D data into field representation to reduce information loss but is still limited to sparse volume.
{\it Multiview CNNs}:
\cite{su2015multi} and \cite{qi2016volumetric} strive to exploit the advances in 2D CNNs by rendering 3D shapes into multi-view images.
Though dominating performance has been achieved on shape retrieval and classification, it remains nontrivial to extend multi-view based approach to other 3D tasks like point classification and shape completion.
{\it Octree/Kd-tree DNNs}:
To provide 3D CNNs the ability to handle high-resolution input, efficient indexing structure like octree and kd-tree are employed in \cite{wang2017cnn, riegler2017octnet} and \cite{klokov2017escape} respectively to reduces the memory cost.
Though being more scalable in parameter sharing than uniform grid, tree-structure based approaches are lacking the overlap of receptive fields between different cells.
{\it Point DNNs}:
PointNet~\cite{charles2017pointnet} is the first work to directly apply deep learning on point set, in which MLPs are employed to extract per-point features, followed by a maxpooling operation to obtain a global feature vector. 
As PointNet lacks the ability to capture local context, PointNet++~\cite{qi2017pointnet++} is later proposed to perform hierarchical feature extraction from grouped points in different levels. 
However, the heuristic grouping and sampling scheme is designed for combining feature from multiple levels and thus fails to explicitly model the spatial point distribution.
More recently, GeoNet~\cite{he2019geonet} proposes to model the intrinsic structure of point cloud based on a novel geodesic-aware representation.
Our approach provides explicit modeling of the spatial distribution of point cloud, which in turn generates more discriminative features for point cloud learning.


\section{Method}
\label{sec:method}


\subsection{Radial Basis Function}
\label{sec:rbf}

\begin{wrapfigure}{r}{0.24\textwidth}
	\begin{center}
		\includegraphics[width=0.24\textwidth]{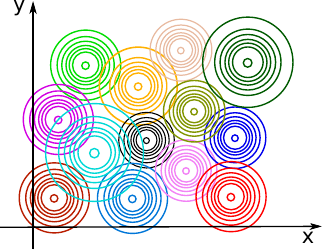}
	\end{center}
	\caption{Overlapping receptive fields.}
	\label{fig:rbf_overlap}
\end{wrapfigure}
Radial basis function is a special class of functions, whose response decreases (or increases) monotonically with distance from its central point.
As its value only depends on input distance, a radial basis function can be generally represented as:
\begin{equation}
	\Phi(\mathbf{x},\mathbf{c}) = \Phi(||\mathbf{x} - \mathbf{c}||),
	\label{eqn:rbf}
\end{equation}

in which $\mathbf{x}$ and $\mathbf{c}$ indicate the input and the center of the RBF kernel respectively. 
The distance metric is usually Euclidean distance (as shown in Equation~\ref{eqn:rbf}) though other metric functions are also possible.

Gaussian function is one of the most commonly used RBF kernels. A typical Gaussian-based RBF function has the following response to the input:
\begin{equation}
    \mathcal{G}(\mathbf{x},\mathbf{c}) = \mathrm{exp}(-\frac{(\mathbf{x}-\mathbf{c})^2}{\sigma^2}).
\end{equation}

Its parameters include the origin $\mathbf{c}$ and the kernel size $\sigma$, which controls the scale of receptive field of a 3D Gaussian kernel when interplaying with neighboring points. 
The Gaussian-like RBF kernels are local given that significant response only falls in the neighborhood close to the origin. 
It is more commonly used than the RBFs with a global response, e.g. a multiquadratic kernel $\Phi(\mathbf{x},\mathbf{c}) = \sqrt{1+\epsilon^2(\mathbf{x}-\mathbf{c})^2}$, where $\epsilon$ is a scaling constant. 
The local kernels are more biologically plausible due to its finite response. 
Other commonly used local kernels include:
Markov function $ \Phi(\mathbf{x},\mathbf{c}) = \mathrm{exp}(-\frac{||(\mathbf{x}-\mathbf{c})||}{\sigma^2})$ and the inverse multiquadratic function $(1+\sigma^2(\mathbf{x}-\mathbf{c})^2)^{-\frac{1}{2}}$.

The localized response of RBF kernels is favorable in describing various spatial distributions. 
In the forward pass of our network, only a sparse set of RBF kernels are highly activated, rendering discriminative patterns for different point distributions.
Yet the overlap between the receptive fields of adjacent kernels provide more diverse response even only sparse points are presented as input. We will demonstrate the robustness of our approach in Section 4.

\subsection{Network Architecture}
\label{sec:network}

\begin{figure*}[htb]
 \begin{center}
  \centering
  \includegraphics[width=0.95\linewidth]{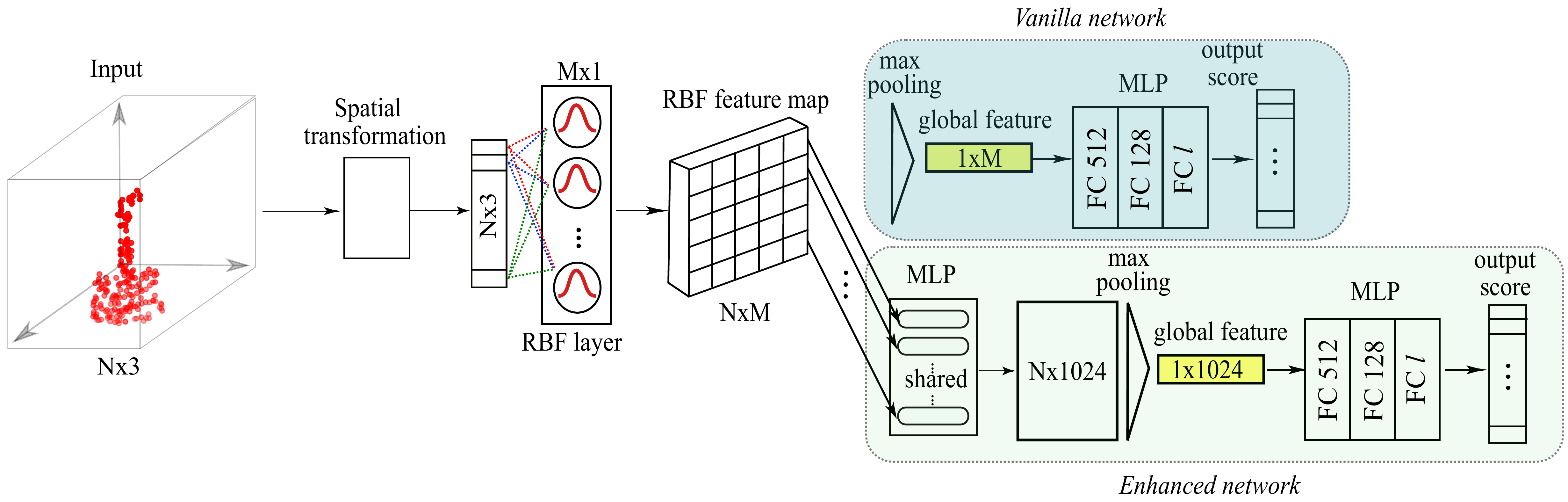}
 \end{center}
 \Caption{Two architectures of our deep RBFNet, including a vanilla version and an enhanced version. $l$ indicates number of classes.}{}
 \label{fig:archi}
 \vspace{-3mm}
\end{figure*}

The architecture of our network is visualized in Figure~\ref{fig:archi}.
We present two network structures: vanilla network and its enhanced version, in which the latter further boosts the performance by introducing a shared MLP layer before the max pooling operation.
The vanilla network contains three key components:
a spatial transform module, a RBF feature extraction layer and a max-pooling layer to aggregate signatures from the neighborhood of each kernel.
We leverage the T-net structure in \cite{charles2017pointnet} for the spatial transform module. 
By introducing a spatial transformer, we expect the input points to be aligned in a canonical space such that the extracted feature is invariant to geometric transformations. 
In particular, it predicts a $3 \times 3$ affine transformation matrix that could be directly applied to the point coordinates.
\vspace{-2mm}

\paragraph{RBF Feature Extraction.}  
At the core of our framework is the RBF feature extraction layer that generates nonlinear signatures given input point cloud.
An RBF layer contains $M$ RBF kernels, each of which has two sets of parameters: the coordinates of the kernel center $c_i$ and the kernel size $\sigma_i$.
Given an input point $x_i$, the RBF layer computes its response from all kernels. Specifically, the activation value for point $x_i$ with respect to kernel $\kappa_j$ can be computed via applying $\kappa_j$'s kernel function $\Phi(x_i, \kappa_j)$.
Therefore, for $N$ input points, the RBF layer will generate a feature map with dimension $N \times M$.

Both the kernel positions and kernel sizes in the RBF layer are optimized so that the resulting kernel configuration is capable to capture the spatial distribution of the input point clouds. 
Compared with fully connected layers, the widely employed component in previous point cloud classification networks, that contain the quantity of parameters equal to the product of the number of neurons in each hidden layer, a single RBF layer can produce more discriminative features with orders of magnitude less parameters. We provide comparisons of performance and time and space complexity with prior works in Section 4.2.
\vspace{-2mm}

\paragraph{Feature Enhancement.} 
In the vanilla structure of our network, the RBF layer is followed by a max pooling operation to extract the most salient point feature for each kernel. 
Thanks to the locality of kernel functions, only a sparse set of kernels will be activated given an input point cloud, generating discriminative features even with such simple structure.
To further enhance the features by leveraging the shared receptive fields between kernels, we extend our network by adding a shared MLP layer before the max pooling. 
The additional MLP contains 3 layers with layer sizes being 16, 128 and 1024 respectively.
The fully connected layers allow the communication between different kernels and can thus further boost the performance of our network. 
The additional MLP layer converts the $N \times M$ feature map into a feature array with dimension $N \times 1024$.
The subsequent max pooling layer then extracts a global feature of fixed length $1024$ for the following linear classifier.
The performance on adding this MLP layer will be detailed in Section 4.

\paragraph{Extension to Additional Channels.}
Our framework is flexible to consider different kernel functions as well as additional channels of input, e.g. local curvatures, point normals, point density and semantic labels.
As the inputs in different channels may have different data distribution, directly concatenating them for feature learning would introduce the difficulty of defining a proper distance metric for measuring the closeness between the input and kernel center.
We therefore propose a multi-channel RBF layer for accommodating multi-channel inputs.
In particular, for each attribute of the input, a separate set/channel of RBF kernels is created and initialized by randomly sampling from the distribution of that attribute.
The feature maps from each RBF channel are then concatenated for further processing in the subsequent layers of network.
To this end, features from different domains are learnt independently via different channels of RBF layers. 
Our experimental results have demonstrated that such strategy of dealing with multi-attribute input could further improve the network performance.

\section{Experiments}
\label{sec:exp}

\subsection{Datasets and Implementation Details}
\label{sec:imp}

As a toy example, the MNIST dataset~\cite{lecun1998gradient} is adopted for the classification task in Section 4.2. 
To ensure the fairness of comparisons, we follow the same procedure proposed in PointNet~\cite{charles2017pointnet} to convert each digit into a 256 two-dimensional points. 
We also evaluate our algorithm on ModelNet40 \cite{wu20153d}, the standard benchmark for the task of 3D object classification. 
ModelNet40 collects 3D models from 40 categories with training and testing sets split into 9843 and 2468 models respectively. 
Models from this dataset have been normalized to $[-1, 1]$ and aligned with a upright orientation.
To simulate the 3D object recognition scenarios in real world, where the "facing" directions of objects are unknown, we augment the data by randomly rotating the shapes horizontally.
We also add random jitterings to the point cloud to approximate the noises generated in real capturing scenarios. 
Note that we apply the data augmentations to both training and testing data. 
Therefore, the orientations of input object are unknown in the testing phase, providing more accurate evaluation of our trained network on the real world data. 

Our network is implemented with Tensorflow~\cite{abadi2016tensorflow} on a NVIDIA GTX1080Ti graphics card.
We optimize the network using Adam optimizer~\cite{kingma2014adam} with an initial learning of 0.0002 and batch size of 32. The learning rate is decreased by $30\%$ for every 20 epochs.
We apply both batch normalization and dropout (keep-probability is set to 0.7) for all the MLP layers.
ReLU is used as the activation for the fully connected layers.
For the results reported in Section 4.2 and Section 4.3, the kernel configurations are randomly initialized unless stated otherwise.
In particular, the kernel origins are initialized by randomly sampling inside a unit sphere centered at $(0,0,0)$.
We draw random samples from a uniform distribution in the range of $[0.01, 1]$ to initialize the kernel sizes.

\subsection{Comparisons}
\label{sec:compare}

\begin{table*}
\centering
\begin{tabular}{l c c | c c c  c c}
\toprule
\multirow{2}{*}{Method} &
\multirow{2}{*}{Input} &
\multirow{2}{*}{Format} &
\multicolumn{3}{c}{ModelNet40} &
\multicolumn{2}{c}{MNIST} \\
 & & & Class & Instance & Time  & Input & Err. Rate \\
\midrule
MVCNN~\cite{su2015multi} & images & 80 views & - & 90.1 & -  & - & - \\
OctNet~\cite{riegler2017octnet} & octree & $128^3$ & 83.8 & 86.5 & - & - & -  \\
O-CNN~\cite{wang2017cnn} & octree & $64^3$ & - & 90.6 & - & - & -  \\
ECC~\cite{simonovsky2017dynamic} & points & $1000 \times 3$ & 83.2 & 87.4 & - & - & 0.63  \\
DeepSets~\cite{zaheer2017deep} & points & $5000 \times 3$ & - & 90.0 & - & - & - \\
Kd-Net~\cite{klokov2017escape} & points & $2^{15} \times 3$ & 88.5 & 91.8 & 120h & $1024 \times 2$ & 0.90 \\
PointNet~\cite{charles2017pointnet} & points & $1024 \times 3$ & 86.2 & 89.2 & 6h & $256 \times 2$ & 0.78 \\
PointNet++~\cite{qi2017pointnet++}  & pts + nor & $5000 \times 6$ & - & 91.9 & 20h & $512 \times 2$ & \textbf{0.51}  \\
\hline
Ours (vanilla) & points & $1024 \times 3$ & 86.3 & 89.1 & \textbf{40min}  & - & -\\
Ours (enhanced) & points & $1024 \times 3$ & 87.8 & 90.2 & 2h   & $256 \times 2$ & 0.58   \\
Ours (enhanced) & pts + nor & $5000 \times 6$ & \textbf{88.8}  & \textbf{92.1} & \textbf{3h}   & - & -  \\
\bottomrule
\end{tabular}
\caption{Results of object classification for methods using different 3D representations.
}
\label{tab:compare}
\end{table*}

We compare the classification accuracy of our algorithm with the state-of-the-art approaches using various representations, e.g. multi-view images, points, kd-tree and octree, which are summarized in Table~\ref{tab:compare}.
Our results reported in Table~\ref{tab:compare} are generated using only 300 RBF kernels.
For the ModelNet40, our vanilla network has achieved comparable performance with PointNet.
However, our approach requires significantly less training time (40 minutes) compared to PointNet (6 hours).
This is due to the nonlinearity introduced by our RBF convolution has enabled remarkably less parameters than the PointNet structure.
By consuming the additional attribute of point normal, the instance classification accuracy of our enhanced model has surpassed the state-of-the-art method - PointNet++ by $0.2\%$ with the same input data size ($5000 \times 6$).
However, unlike PointNet++ which requires 20 hours to converge, our network only needs 3 hours for training.
In the MNIST dataset, our enhanced network outperforms PointNet by $23\%$ in terms of instance accuracy.
Though PointNet++ still performs sightly better than the enhanced model in MNIST classification, our approach can still achieve comparable result using half of the points.

\begin{table}
\centering
\begin{tabular}{l c c c }
\toprule
& $\#$params  &  FLOPs/sample  & Inf. Time   \\
\midrule
PointNet & 3.5M  & 440M  & 1 ms  \\
PointNet++   &  12.4M  & 1467M  & 4 ms  \\
Subvol.~\cite{qi2016volumetric} & 16.6M & 3633M & - \\
MVCNN~\cite{su2015multi} & 60.0M & 62057M & - \\
ours(val.) &  \textbf{2.2M}  & \textbf{24M}  & \textbf{0.09 ms}   \\
ours(enh.) &  3.2M  & 218M  & 0.4 ms   \\
\bottomrule
\end{tabular}
\vspace{2mm}
\caption{Comparisons of time and space complexity. FLOP stands for floating-point operation. "M" stands for million while "ms" represents millisecond.}
\label{tab:time}
\end{table}

\paragraph{Time and Space Complexity Analysis.} 
Table~\ref{tab:time} summarizes the time and space complexity of our networks and a representative set of related approaches, such as point cloud, volumetric \cite{qi2016volumetric} and multi-view~\cite{su2015multi} based architectures. 
The time complexity is measured by FLOPs and inference time per sample (column 2 and 3). Our enhanced model is orders more efficient in computational cost: \textit{284x}, \textit{16x} and \textit{6.7x} less FLOPs per sample than MVCNN~\cite{su2015multi}, Subvolume~\cite{qi2016volumetric} and PointNet++ respectively.
In terms of inference time, our vanilla and enhanced network are \textit{44x} and \textit{10x} faster than PointNet++.
Regarding space complexity, our enhanced network uses \textit{20x}, \textit{5x} and \textit{4x} less parameters compared to MVCNN, Subvolume and PointNet++ respectively.

\subsection{Ablation Analysis}
\label{sec:ablation}

To further evaluate the effectiveness and robustness of our algorithm, we provide an ablation analysis using different configurations.
All the following evaluations are performed using Gaussian kernel, 1024 points as input and 300 kernels for feature extraction unless otherwise stated.



\paragraph{Effect of Adding an RBF Layer.}
Our network is extremely simple compared to prior frameworks for point cloud classification.
If not considering the spatial transform module, the introduction of RBF layer is the only difference between our vanilla network and the classic MLP classifier.
Therefore, a natural question would be how the RBF layer could improve the performance for point cloud classification.
In our experiment, a MLP layer with only a spatial transformer can achieve nearly $100\%$ accuracy in training set but performs extremely poor on the test set ($4.5\%$ accuracy), indicating a strong overfitting to the training data.
However, by introducing a single RBF layer with 300 kernels, the performance is improved dramatically to the accuracy of $89.1\%$.
The significant improvement of performance has proven the strong competency of RBF layer in generating discriminative features for understanding 3D point clouds.

\begin{figure*}[tbh]
	\begin{center}
		\subfloat[]{
			\label{fig:numKernel}
			\includegraphics[width=0.26\linewidth]{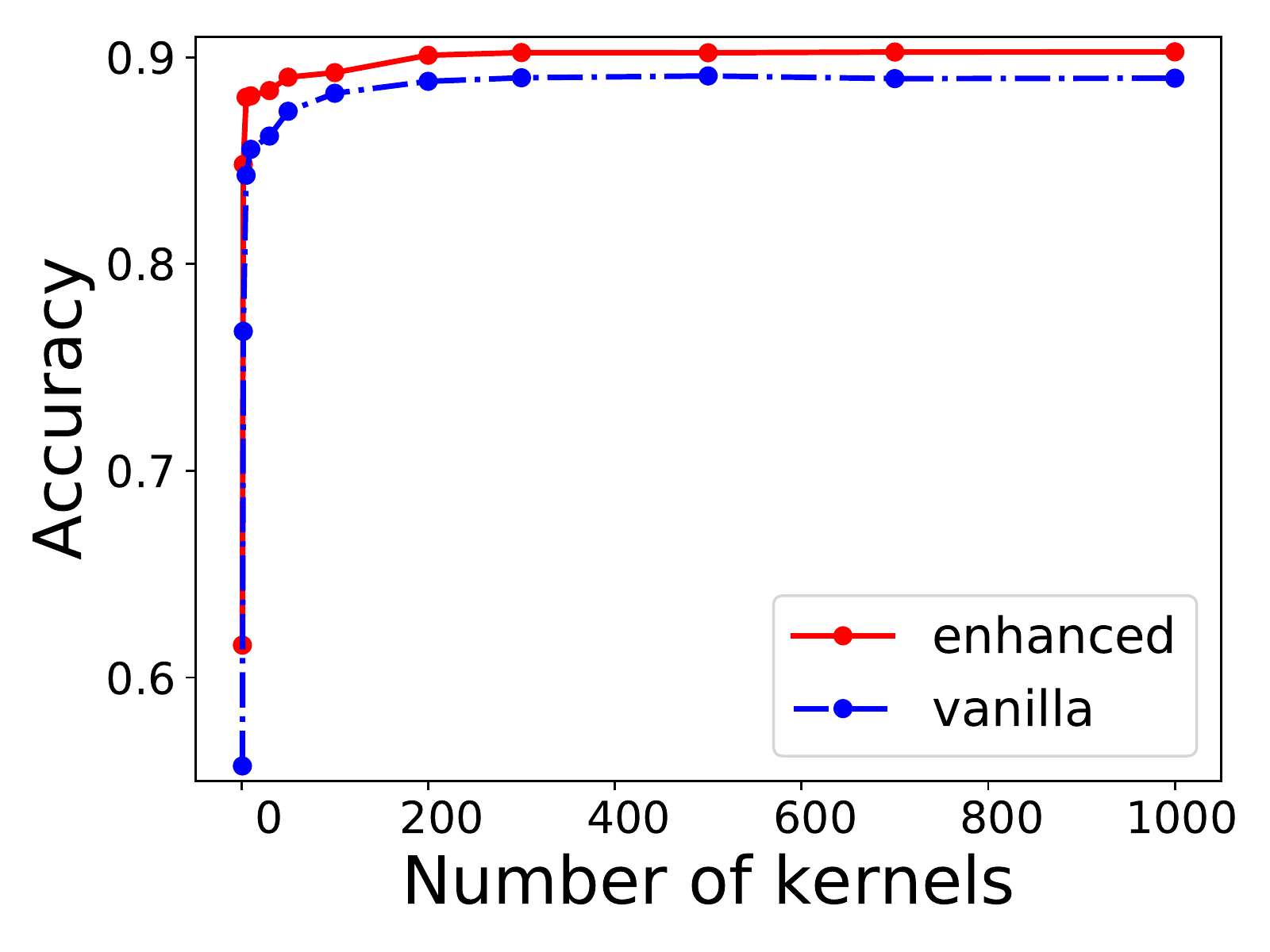}}
		\subfloat[]{
			\label{fig:numPoint}
			\includegraphics[width=0.26\linewidth]{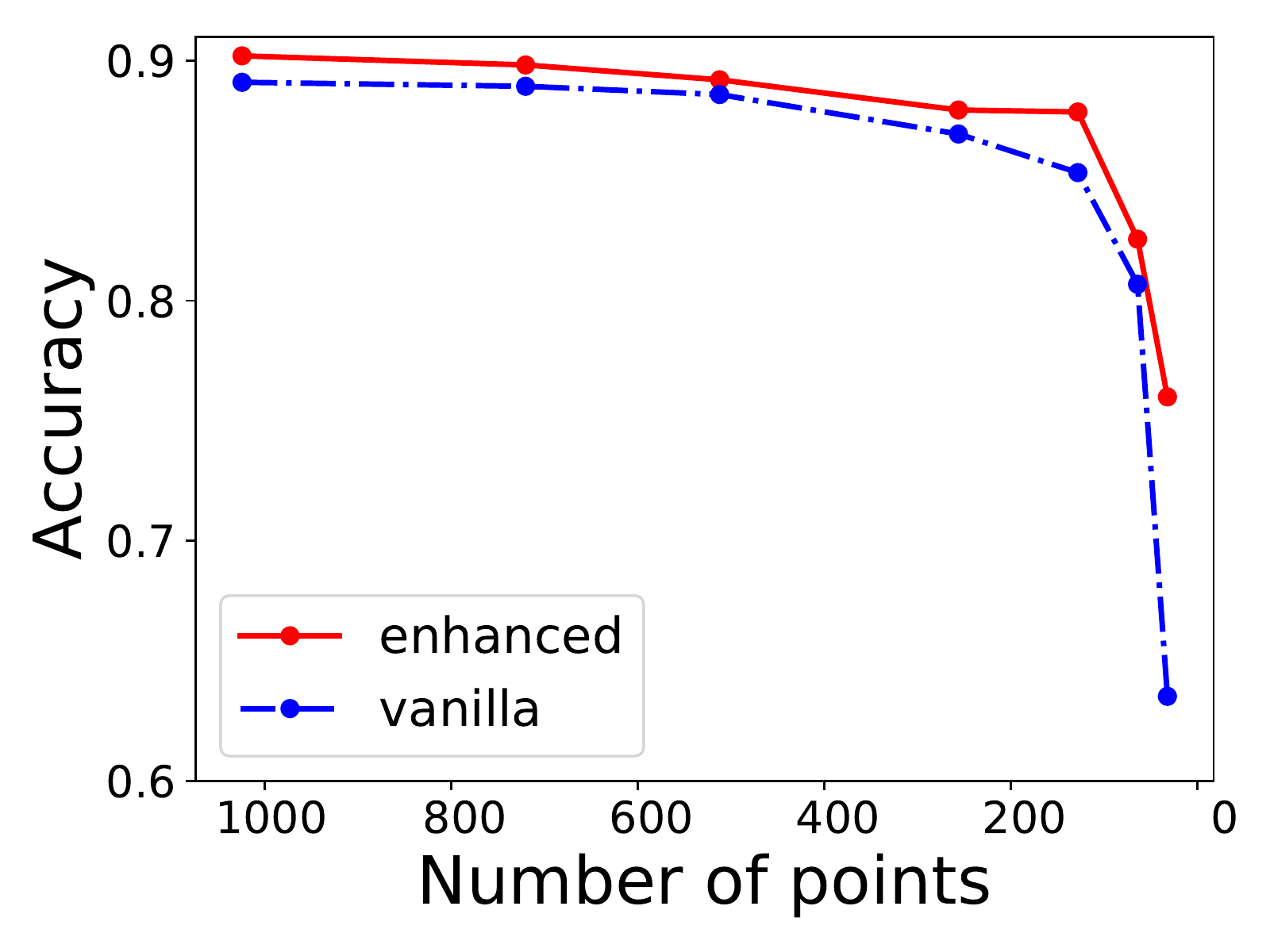}}
		\subfloat[]{
		\label{fig:5Kernel}
		\includegraphics[width=0.20\linewidth]{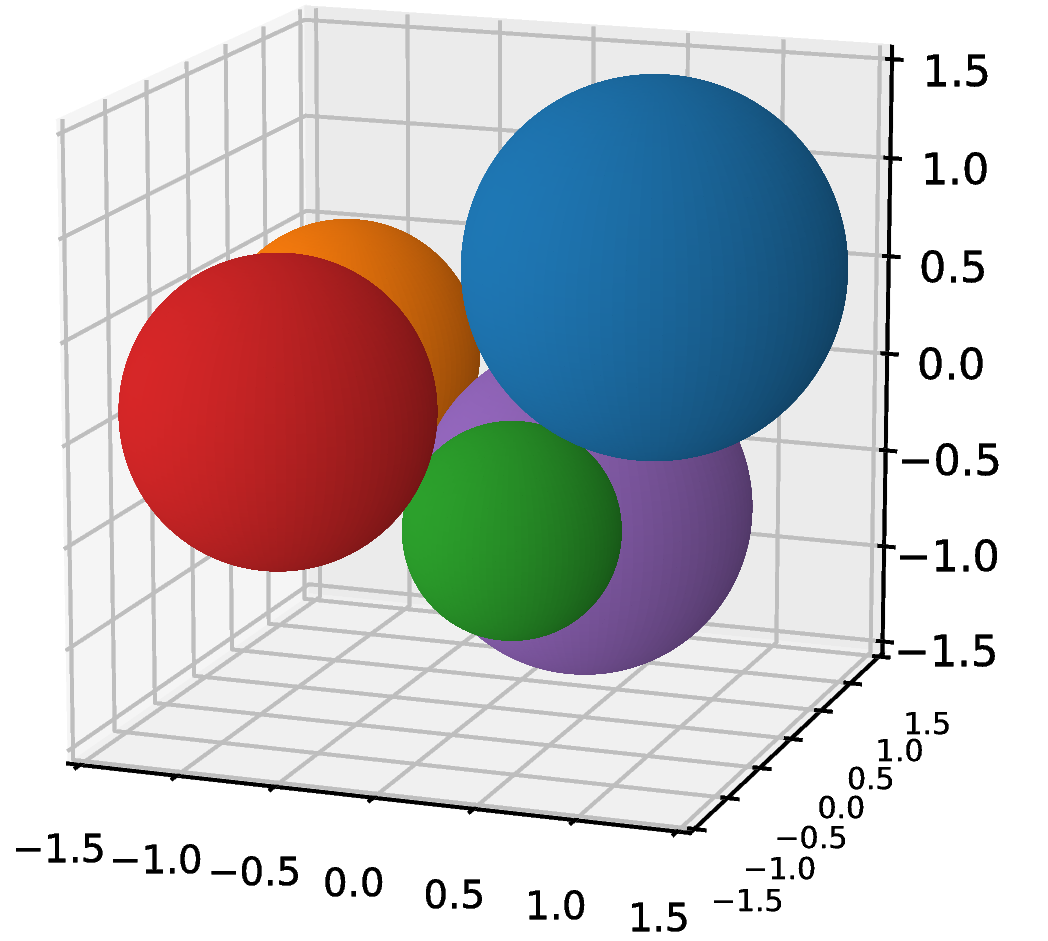}}
		\subfloat[]{
			\label{fig:300Kernel}
			\includegraphics[width=0.21\linewidth]{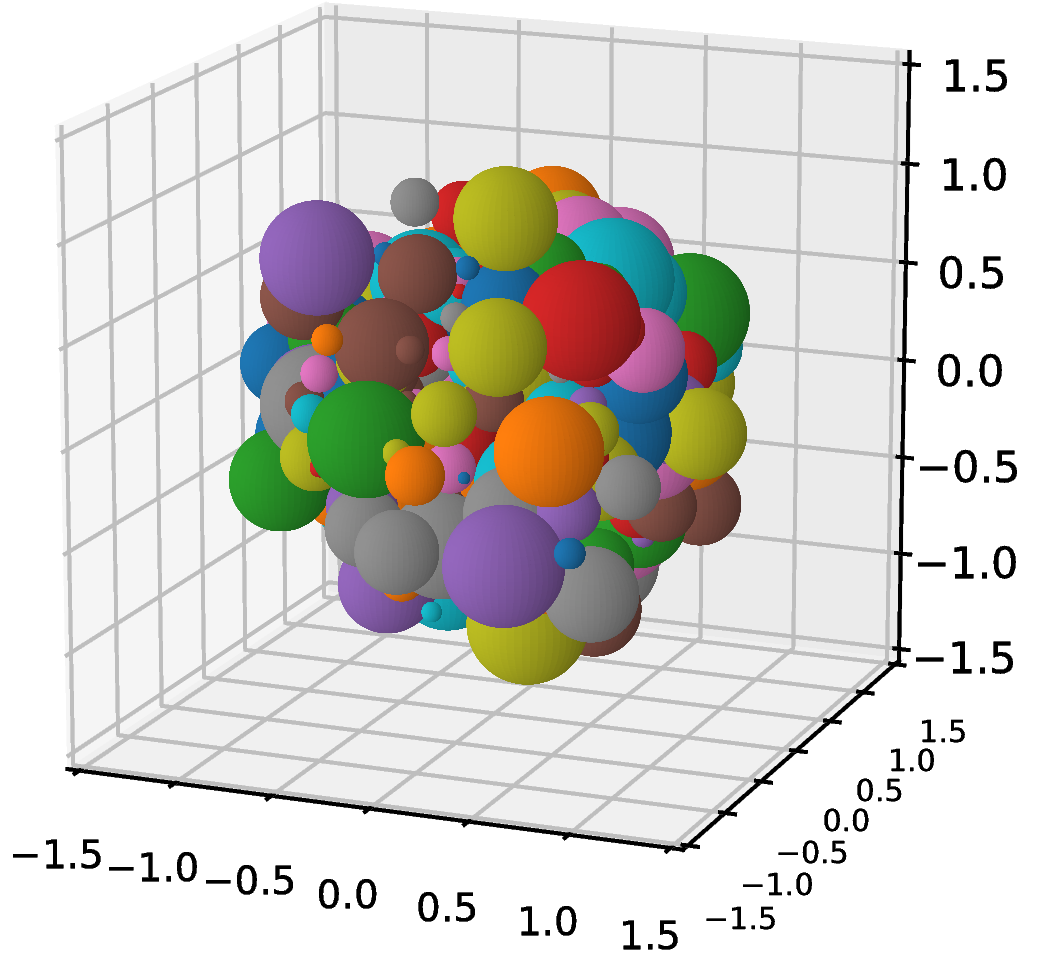}}
	\end{center}
	\caption{Accuracy curve with the corruption of (a) kernel number  and (b) number of input points. (c) and (d) visualize the kernel configurations when 5 and 300 kernels are used respectively. Each kernel is displayed as a sphere, whose origin and radius decode the kernel center and size respectively.
	}
	\label{fig:ablation}
\end{figure*}

\begin{table}
\centering
\begin{tabular}{l c c c | c c}
\toprule
& -$25\%$  &  -$50\%$  & -$75\%$  & 0.05 & 0.1 \\
\midrule
PointNet & 86.1  & 83.2  & 74.0  & 79.1 & 30.0 \\
PointNet++   &  89.5  & 87.9  & 86.5  & 82.3 & 45.1  \\
ours(val.) &  88.9  & 88.5  & 86.9   & 85.2 & 62.4  \\
ours(enh.) &  \textbf{89.9}  & \textbf{89.2}  & \textbf{87.9}   & \textbf{86.3} & \textbf{65.6} \\
\bottomrule
\end{tabular}
\vspace{2mm}
\caption{Comparisons of classification accuracy on ModelNet 40. Left 3 columns: results with random point dropout; right 2 columns: results with different Gaussian noise std..}
\label{tab:robustness}
\end{table}


\paragraph{Robustness to Noise and Incomplete Data.}
In the real-world capturing scenarios, the data obtained from sensors is prone to be noisy and highly irregular.
To validate the robustness of our network to the non-uniform and sparse data, we train our network using 1024 points and test it with random point dropout.
Starting from 1024 points, we gradually decrease the amount of input points.
As shown in Figure~\ref{fig:numPoint}, when half of the points (512) are missing, the accuracy of both the vanilla and enhanced network only drop by less than $1\%$.
Our network can achieve more than $85\%$ accuracy when only 128 points are used: vanilla network - $85.3\%$; enhanced model - $87.8\%$.
Our approach is scalable to extremely sparse points.
In particular, when only 32 input points are presented, our enhanced network can still achieve $75.3\%$ accuracy while the vanilla model reaches to $63.5\%$ accuracy.

We also compare the performance of our network with PointNet and PointNet++ when missing points and noises are present (see Table~\ref{tab:robustness}). In the left part of Table~\ref{tab:robustness} (left 3 columns), we compare the classification accuracy in terms of varying degrees of random point dropout. The results demonstrate that our approaches are much more robust than PointNet based networks especially when the input point cloud becomes sparser.
In the right two columns of Table~\ref{tab:robustness}, We compare the results when Gaussian noise is added to each point independently. In particular, we test the candidate approaches with the standard variation of Gaussian noises being 0.05 and 0.1 respectively. 
Our solution significantly outperforms the PointNet structures. 
As shown in Table~\ref{tab:robustness}, our vanilla network outperforms PointNet++ by more than $17\%$ when the input point cloud is highly noisy (standard deviation equals to 0.1). 

\paragraph{Robustness to Kernel Density.}
As the nonlinear RBF kernels employed in our framework only have localized response to the input points, one of our major concerns is whether the network scales well to a sparse set of kernels.
We test our approach with number of kernels ranging from 1 to 1000 as demonstrated in Figure~\ref{fig:numKernel}.
To our surprise, the network works quite well even with two RBF kernels - the vanilla and enhanced structure have achieved $76.7\%$ and $84.8\%$ accuracy respectively.
With five kernels, both networks have obtained accuracy higher than $84\%$: vanilla - $84.3\%$; enhanced - $88.1\%$.
When the number of kernels is further increased, the classification accuracy grows slowly and achieves its maximum value at around 300.
We visualize the optimized kernel distribution when 5 and 300 kernels are used for feature extraction in Figure~\ref{fig:5Kernel} and Figure~\ref{fig:300Kernel} respectively.
As indicated in the visualization, our method automatically increases the receptive field of kernels when kernel density is small, enabling a faithful capture of the spatial point distribution even when extremely sparse kernels are employed.

\paragraph{Variations of Kernel Functions}

\begin{table}
\centering
\begin{tabular}{l c c c c c}
\toprule
 & Gs.  &  Mk. & IMQ & Gs.+Mk. & Gs.+IMQ   \\
\midrule
Vanilla    & 89.1   & 89.0  & 85.6 & 89.5 & 88.3      \\
Enhan.   &  90.2  & 90.1  & 89.8 & 90.5 & 90.3      \\
\bottomrule
\end{tabular}
\vspace{2mm}
\caption{Instance classification accuracy using various kernel functions and their combinations. "Gs." stands for Gaussian while "Mk." stands for Markov.}
\label{tab:kernel_type}
\end{table}

Our framework scales well to different types of kernel functions.
We investigate the performance of our algorithm by using three major types of RBF kernels: Gaussian, Markov and Inverse Multiquadratic (IMQ).
The three candidate kernel functions all have localized activation but differ in the degrees of locality.
In general, Gaussian kernel has the strongest localized response, followed by the Markov function and the IMQ kernel.
As demonstrated in Table~\ref{tab:kernel_type}, Gaussian kernel has achieved the best classification accuracy in both versions of networks.
The performance of Markov function is comparable to that of Gaussian while the accuracy of IMQ is significantly lower than the other two with the vanilla network.
This suggests that more localized/nonlinear response has stronger capability to extract more discriminative features.
The enhanced structure, however, has achieved high accuracy with all the three kernels.
The additional MLP layer has boosted the communication between the kernels, generating more effective features while providing stronger robustness to the variations of learnt features.

Our approach is flexible to consider different kernels.
In particular, similar to considering additional input attributes , multiple channels of RBF layers are employed in which each channel employs a unique kernel.
The feature maps obtained from different channels are concatenated after the RBF layer for further processing.
As indicated in Table~\ref{tab:kernel_type}, it is interesting to observe that the mixture of Gaussian and the other two kernels can produce even higher accuracy than applying them alone.

\begin{table}
	\centering
	\begin{tabular}{l c c c c}
		\toprule
		& Fix Center  &  Fix Size   & Fix Both & Optim. Both   \\
		\midrule
		Vanilla    & 88.6   & 75.2  & 72.9 & 89.1     \\
		Enhan.   &  89.5  & 87.5  & 86.8 & 90.2       \\
		\bottomrule
	\end{tabular}
\vspace{2mm}
	\caption{Effect of optimizing center position and receptive fields of kernels. Instance classification accuracy (in percentage) is shown in the table.}
	\label{tab:fix_optim}
\end{table}

\paragraph{Importance of Optimizing Kernel Center and Nonlinearity.}
\nothing{
\begin{itemize}
    \item Fixed kernel position and size
    \item Fixed kernel position + optimized kernel size
    \item Fixed kernel size + optimized kernel position
    \item Ours (both optimized)
\end{itemize}
}
The center position and kernel size are the two important parameters that completely determine the configuration of RBF layer.
We investigate the importance of each parameter by selectively fixing one of the components.
Table~\ref{tab:fix_optim} demonstrates the instance classification accuracy of four combinatorial settings.
All the settings share the same kernel initialization.
As seen from the Table~\ref{tab:fix_optim}, optimizing both parameters have led to the best performance while fixing both generates the worst results.
If only optimizing kernel size (fixing kernel centers),
the accuracy of both networks only drop by less than one percent.
However, in the case of fixing the kernel sizes, the performance of both networks degrades significantly, especially the vanilla structure.
The experimental results indicate that an optimized nonlinearity of the RBF kernels is the key to extracting discriminative features.
As the initial kernel centers are randomly dispersed inside the unit sphere, the kernels have been able to cover features from each local regions via adjusting its receptive field.
Hence, the optimization of kernel centers has less impact on the final classification accuracy.


\begin{figure*}[tbh]
	\begin{center}
		\subfloat[]{
			\label{fig:uniformb}
			\includegraphics[width=0.18\linewidth]{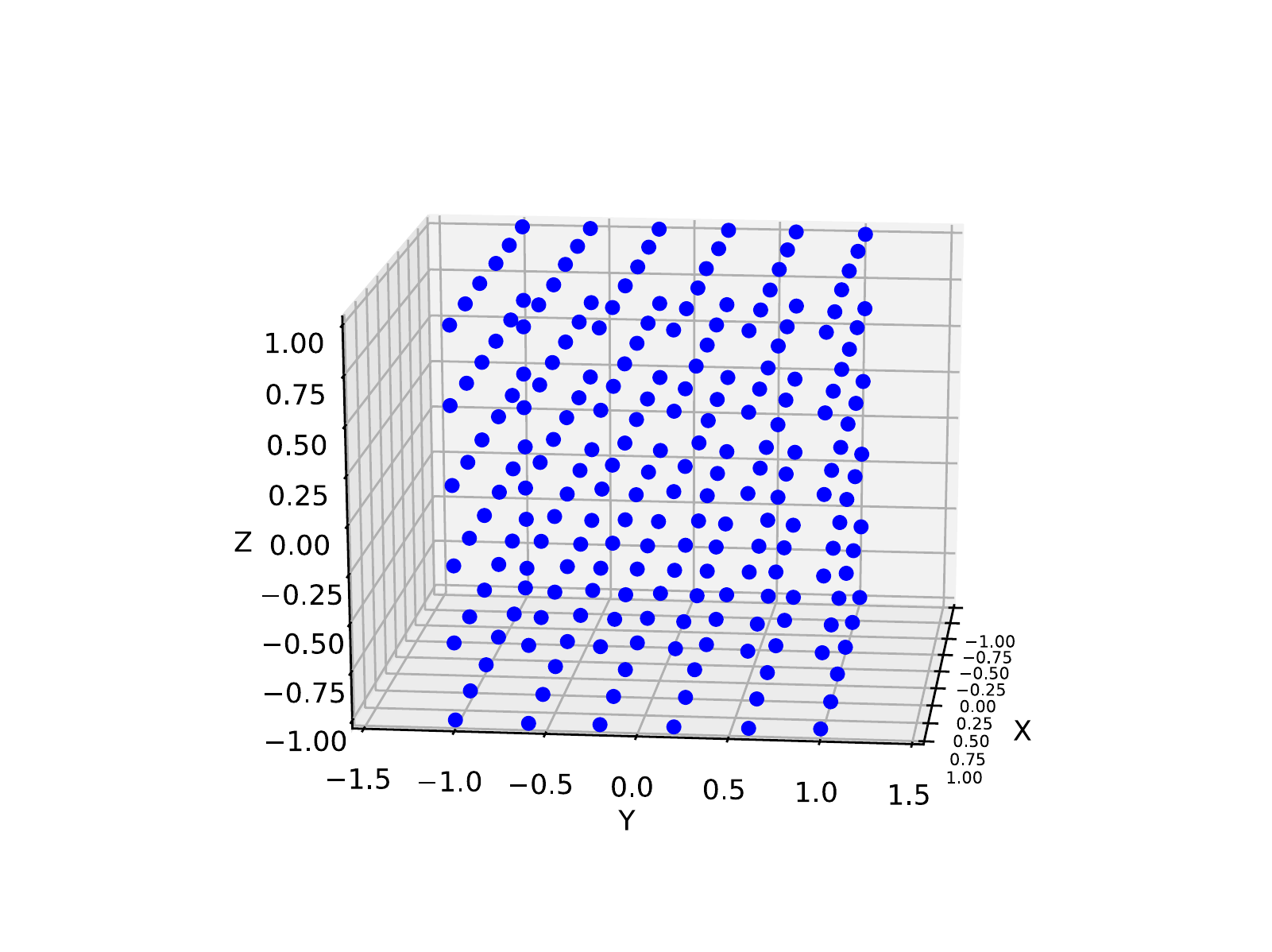}}
		\subfloat[]{
			\label{fig:randomb}
			\includegraphics[width=0.18\linewidth]{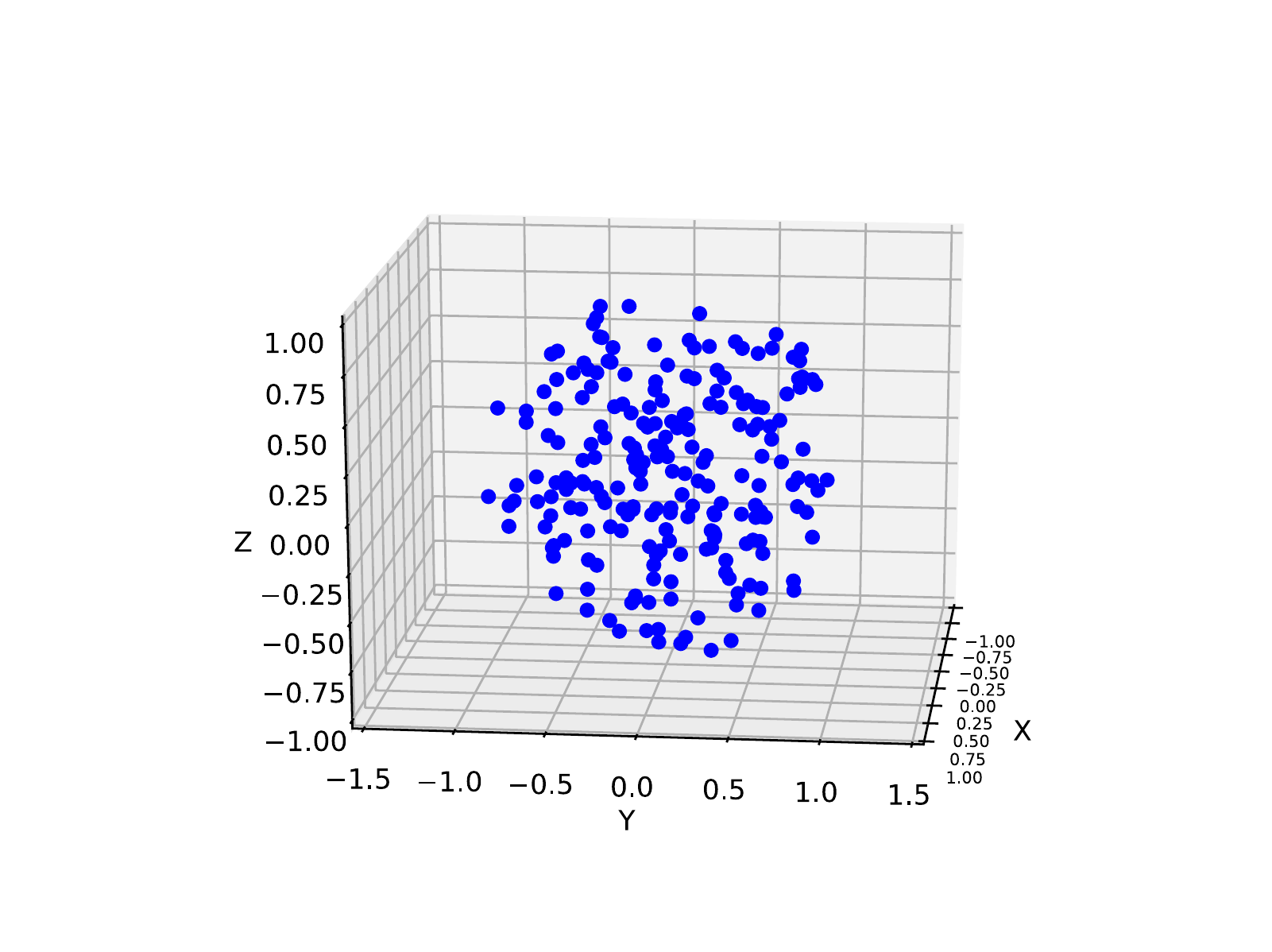}}
		\subfloat[]{
			\label{fig:kmeansb}
			\includegraphics[width=0.18\linewidth]{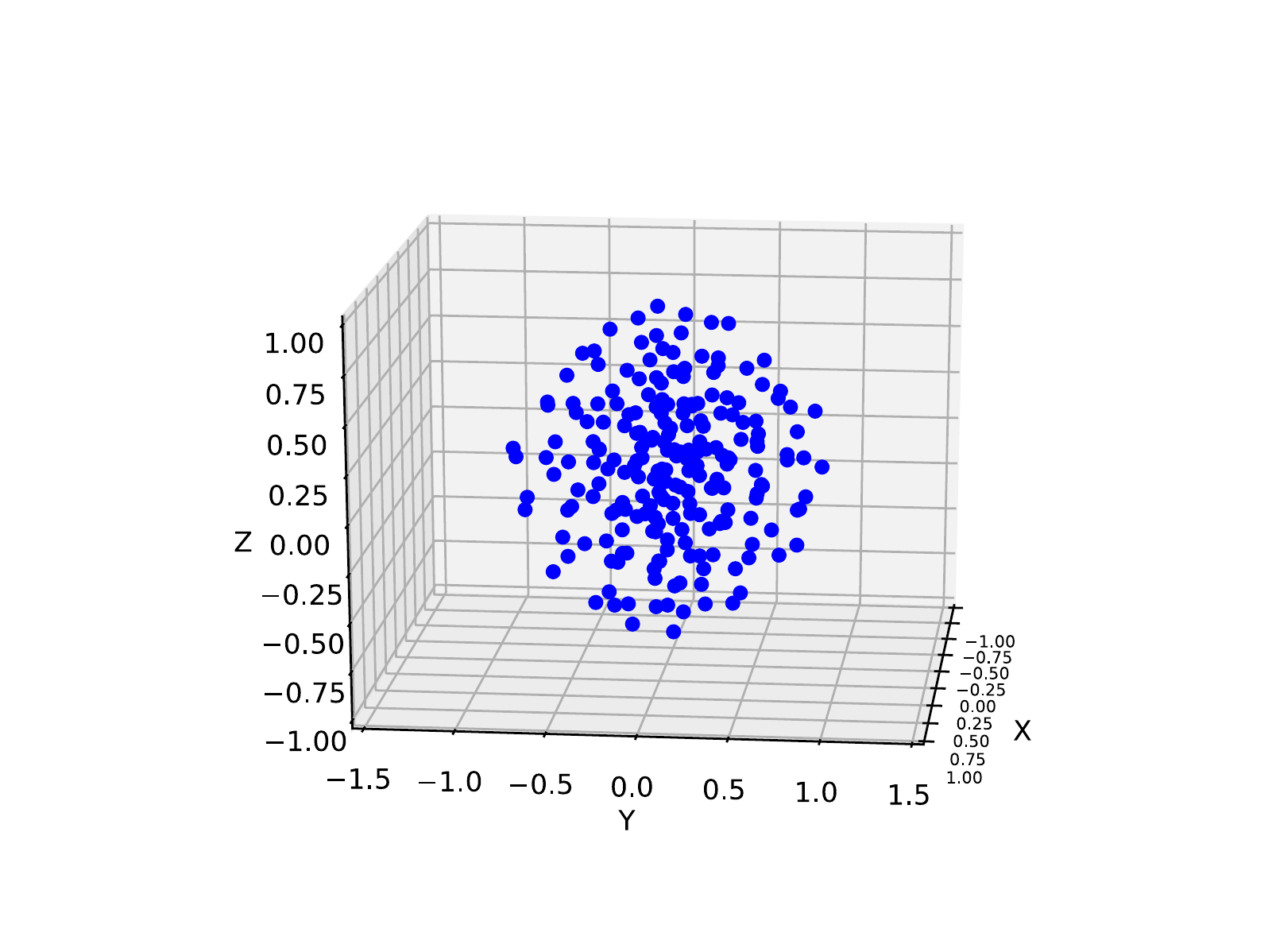}}
			\subfloat[]{
			\label{fig:overlapb}
			\includegraphics[width=0.18\linewidth]{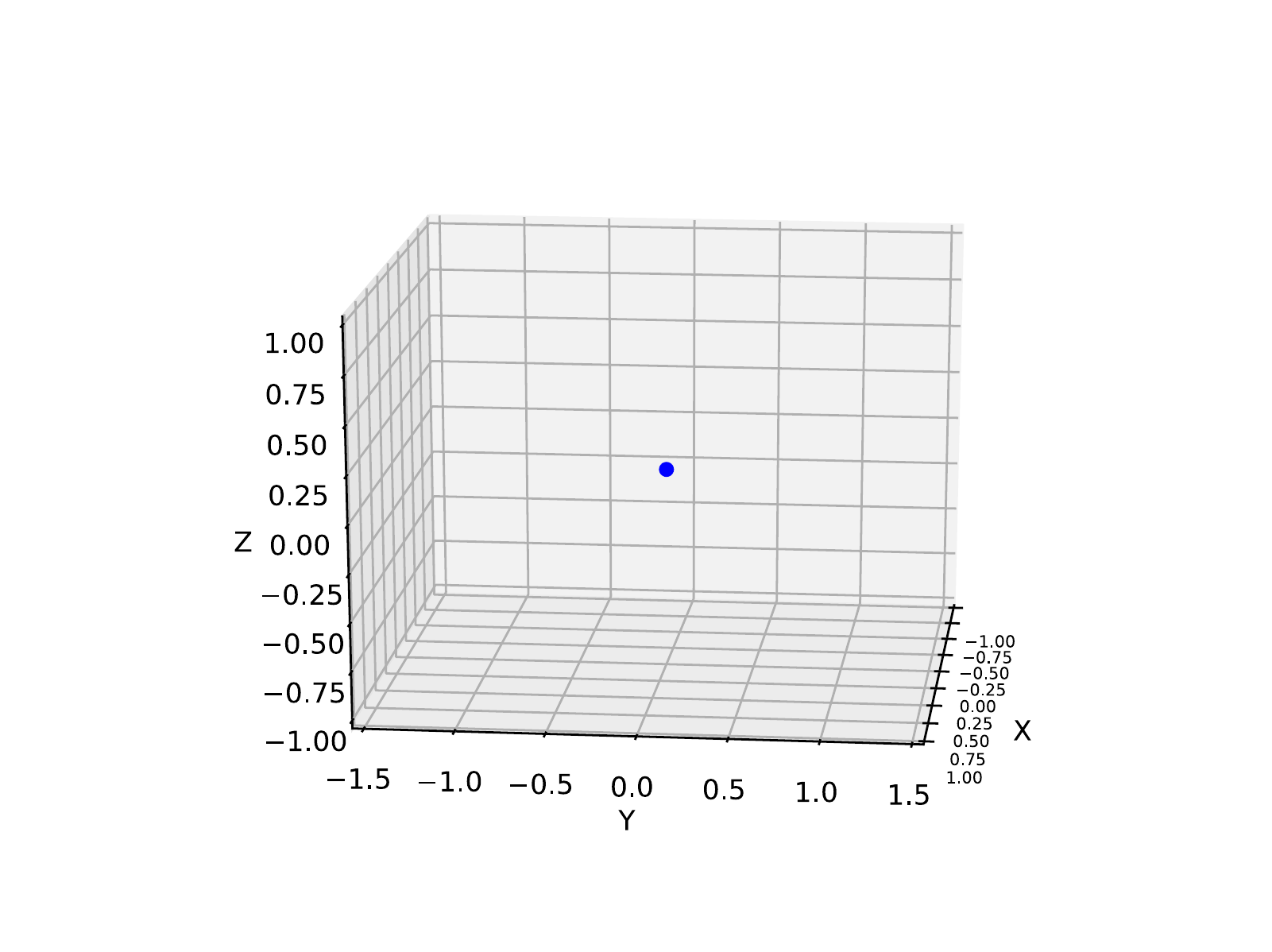}}
		\subfloat[]{
		\label{fig:localb}
		\includegraphics[width=0.18\linewidth]{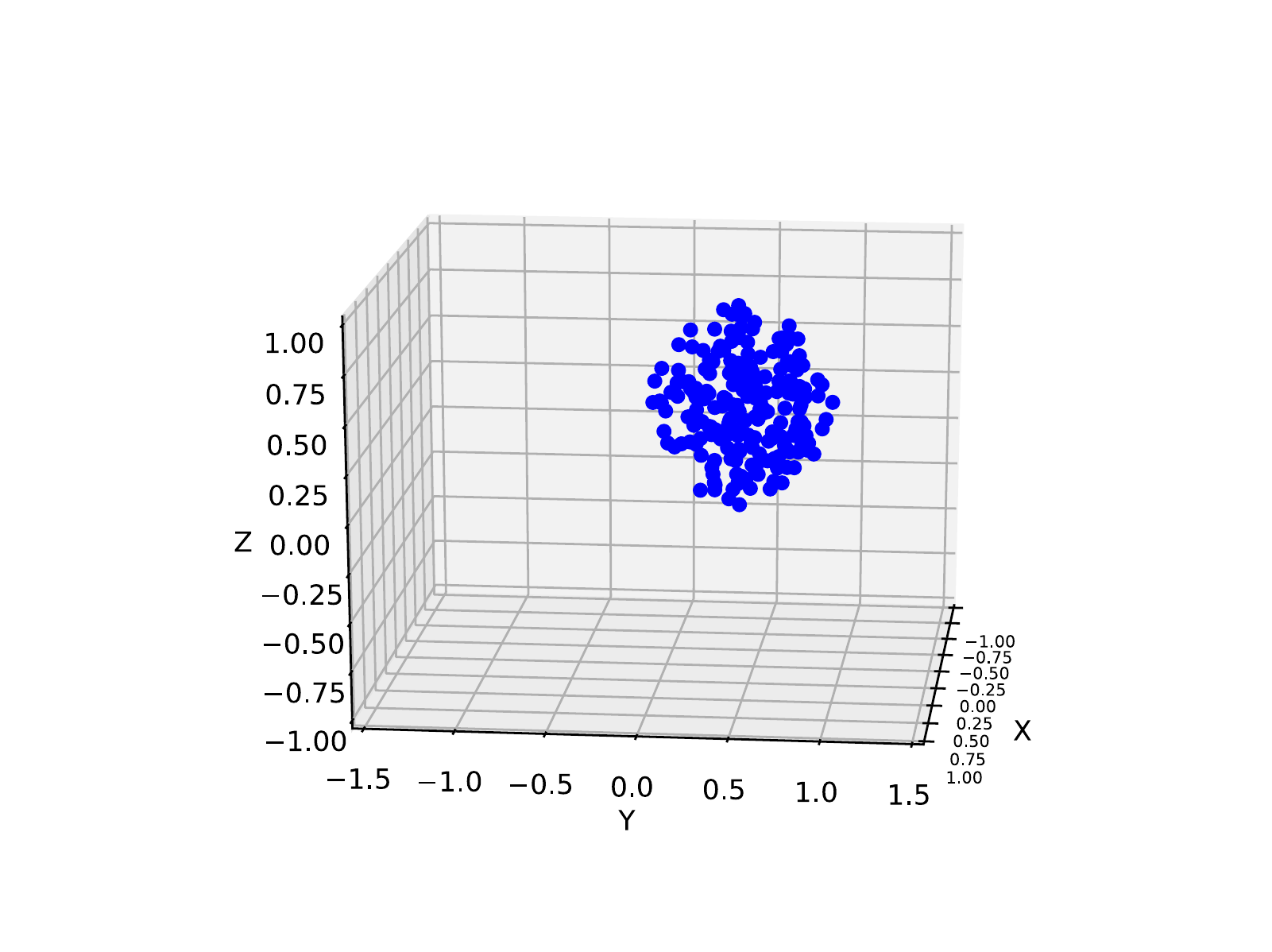}}
		
		\subfloat[Uniform]{
			\label{fig:uniforma}
			\includegraphics[width=0.18\linewidth]{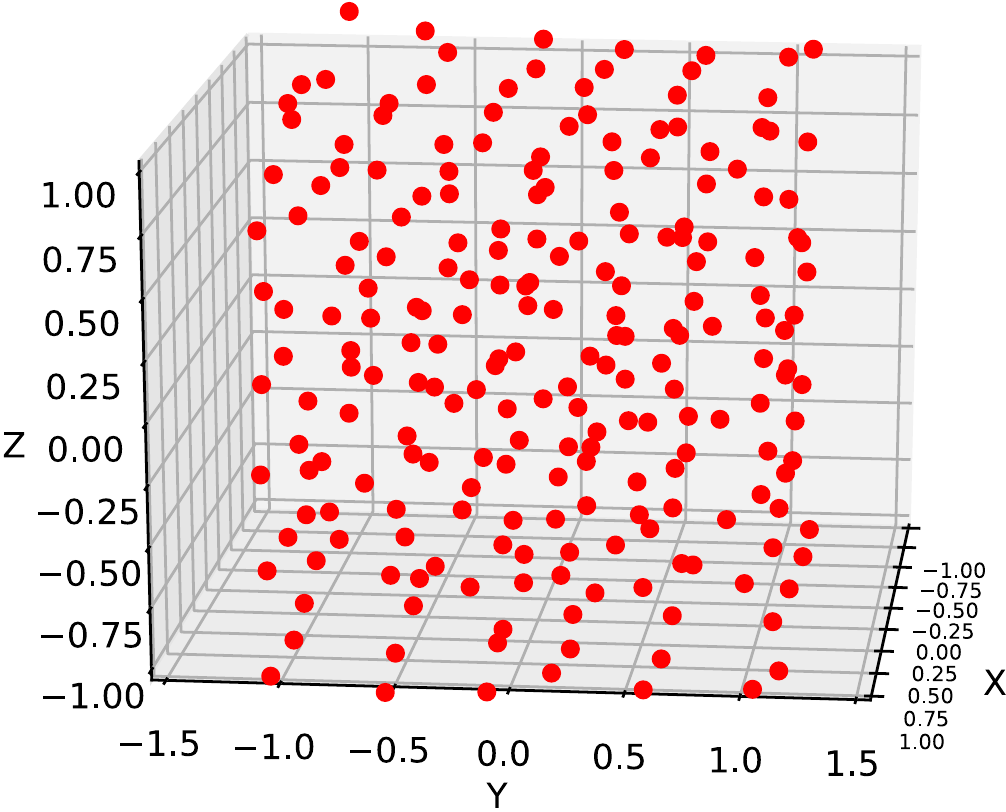}}
		\subfloat[Random]{
			\label{fig:randoma}
			\includegraphics[width=0.18\linewidth]{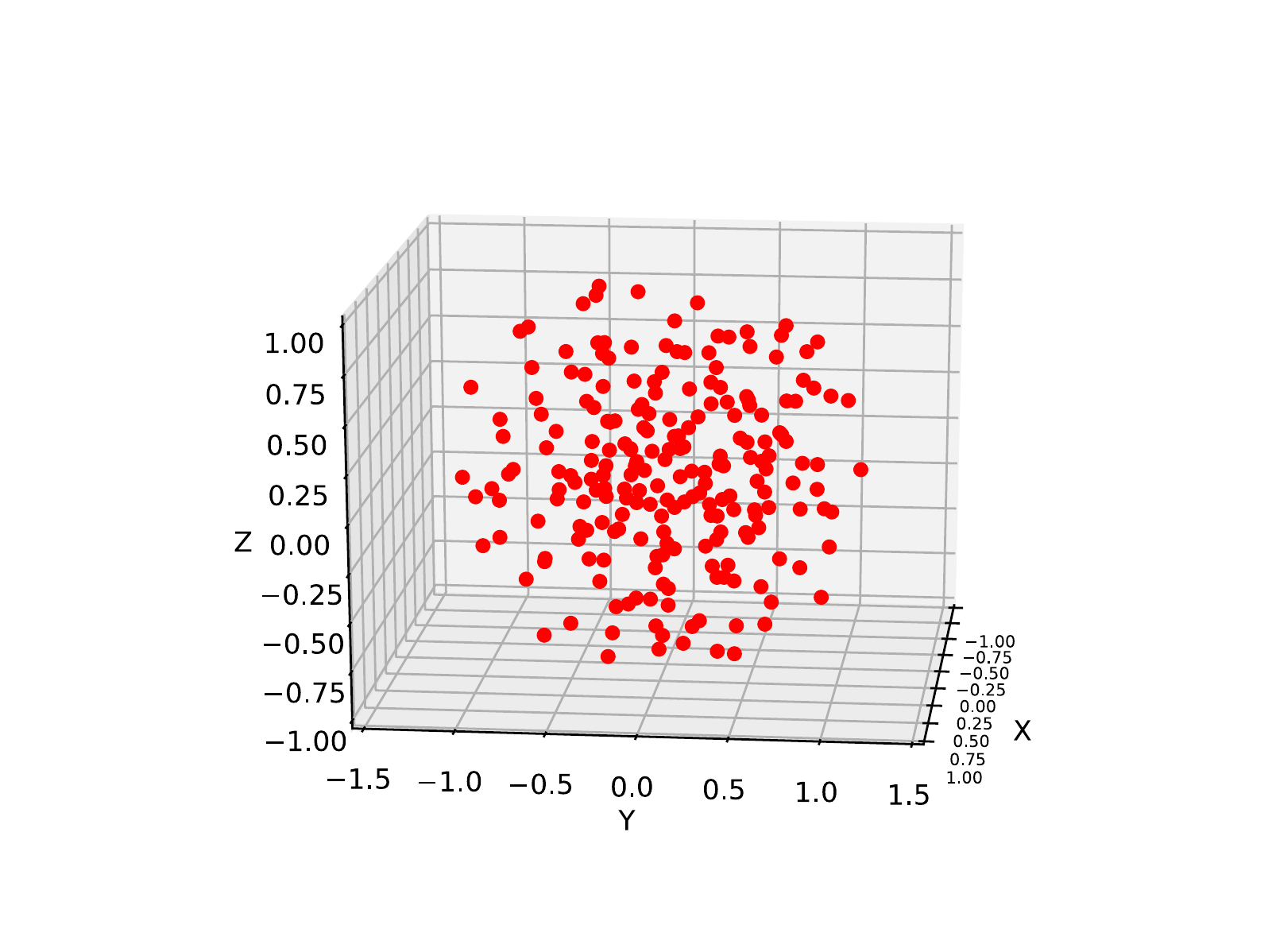}}
		\subfloat[K-means]{
			\label{fig:kmeansa}
			\includegraphics[width=0.18\linewidth]{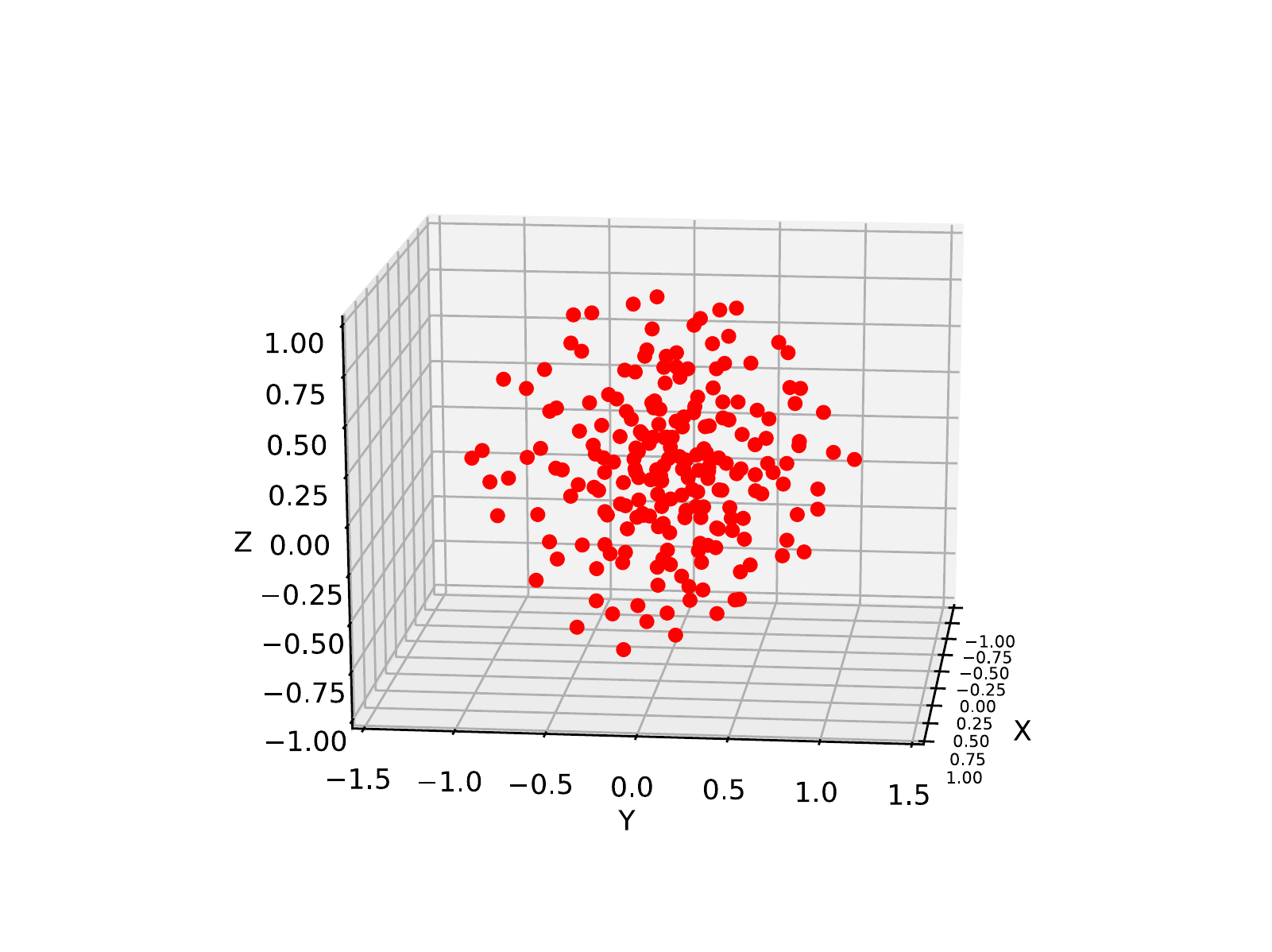}}
			\subfloat[Overlap]{
			\label{fig:overlapa}
			\includegraphics[width=0.18\linewidth]{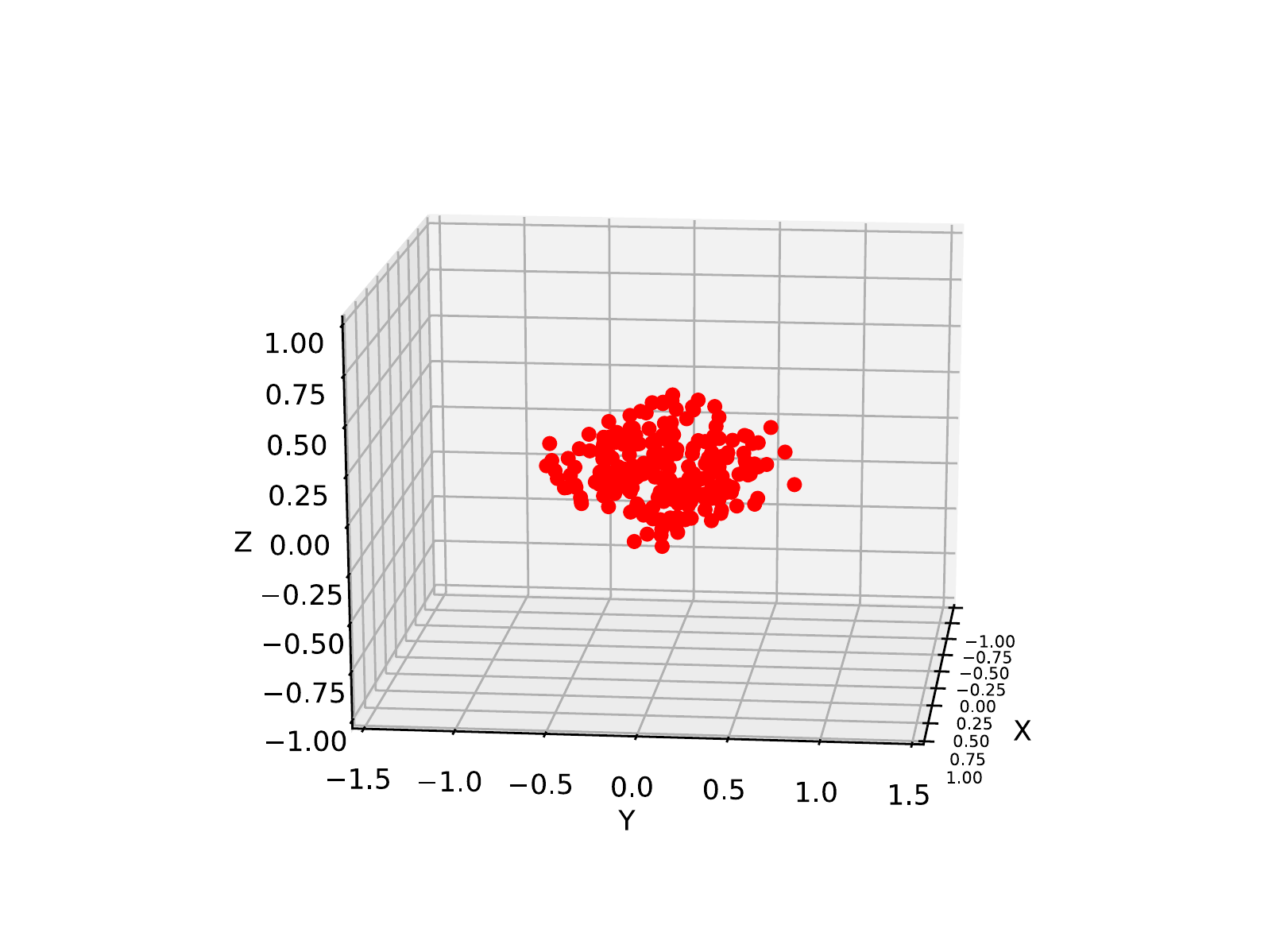}}
		\subfloat[Local]{
		\label{fig:locala}
		\includegraphics[width=0.18\linewidth]{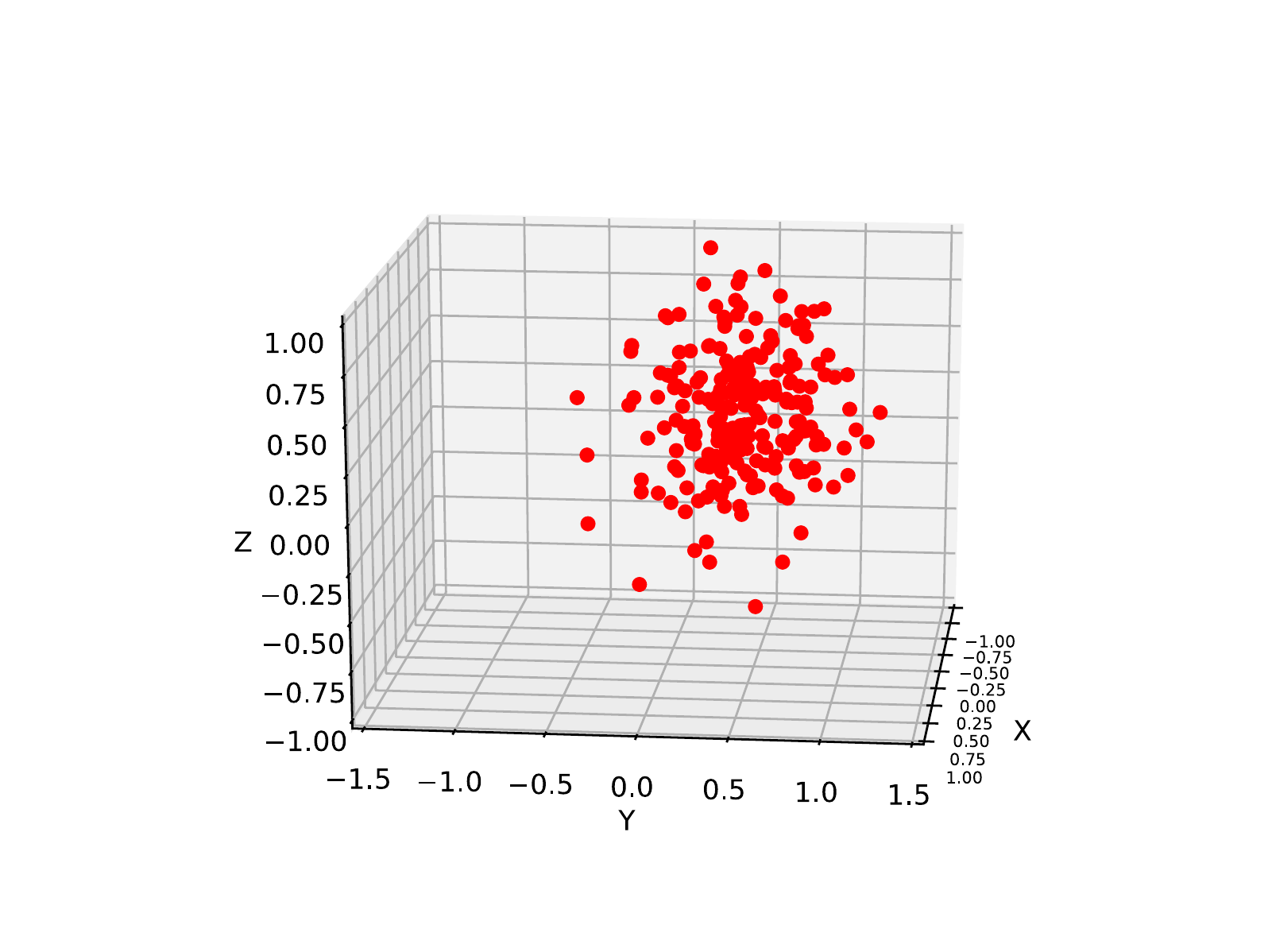}}
	\end{center}
	\vspace{-3mm}
	\caption{Visualization of kernel distributions before (first row) and after (second row) optimization.
	}
	\label{fig:vis}
\end{figure*}


\paragraph{Robustness to Initial Kernel Distribution.}

As the capability to capture the spatial point distribution is the key to analyzing point clouds, we further validate the robustness of our algorithm to various initial kernel distributions.
We investigate five different scenarios: uniform distribution, random sampling within a unit sphere, K-means clustering,
overlapping kernels (all kernels are placed in $(0,0,0)$) and random sampling from a local region.
Figure~\ref{fig:vis} shows the kernel distribution before and after optimization using the vanilla network.
As seen from the results, our network strives to scatter the initial points so as to capture features in a larger scale, especially when the initial sampling is too crowded (see Figure~\ref{fig:overlapb} and Figure~\ref{fig:overlapa}).
Hence our network is still capable to maintain high classification accuracy when the initialization is strongly biased ($88.7\%$ for kernel overlapping; $89.2\%$ for localized initialization).
As seen from Table~\ref{tab:kernel_init}, our approach is robust to different patterns of initialization as long as kernels are dispersed in a global scale.

\begin{table}[h!]
	\centering
	\begin{tabular}{l c c c c c}
		\toprule
		& Uniform  &  Rand.  & K-means & Overlap & Local   \\
		\midrule
		Vanilla & 89.2  & 89.1  & 89.0 & 86.3 & 88.2           \\
		Enhan.   &  90.3  & 90.2  & 90.2 & 88.7 & 89.2      \\
		\bottomrule
	\end{tabular}
	\vspace{2mm}
	\caption{Instance classification accuracy given different initial kernel initializations.}
	\label{tab:kernel_init}
\end{table}


\nothing{
\begin{itemize}
    \item optimized kernel distribution when initial kernels are all overlapped
    \item kernel positions and size before and after optimization
\end{itemize}
}

\section{Conclusion}
\label{sec:conclusion}

In this paper, we present deep RBFNet - a novel deep learning framework that utilizes nonlinear mapping provided by RBF kernels for efficient learning of point cloud features.
Unlike prior works that obtain point features in an implicit manner, we explicitly model the spatial distribution of point cloud by leveraging the localized nonlinearity of RBF kernels.
The local receptive field of RBF kernels enables discriminative patterns of activated kernels for different point distributions.
Thanks to the effective feature extraction, our approach has achieved comparable or even better performance in the task of point cloud recognition comparing to the state-of-the-art methods.
Furthermore, our algorithm has reduced the number of network parameters by a considerable amount, enabling a remarkably faster training than existing works.
We also provide intuitive analysis towards understanding of our networks.

\bibliographystyle{unsrt}
\bibliography{nips2018}

\ifthenelse{\equal{\final}{0}}
{
}
{}
\end{document}